\pdfoutput=1
\documentclass[10pt,twocolumn,letterpaper]{article}

\usepackage{cvpr}
\usepackage{times}
\usepackage{epsfig}
\usepackage{graphicx}
\usepackage{amsmath}
\usepackage{amssymb}
\usepackage{float}
\usepackage{amsmath,diagbox,tabu,stackengine}
\usepackage[utf8]{inputenc} % allow utf-8 input
\usepackage[T1]{fontenc}    % use 8-bit T1 fonts
\usepackage{url}            % simple URL typesetting
\usepackage{booktabs}       % professional-quality tables
\usepackage{amsfonts}       % blackboard math symbols
\usepackage{nicefrac}       % compact symbols for 1/2, etc.
\usepackage{microtype}      % microtypography
\usepackage{authblk}
\usepackage{caption}
\usepackage{multirow}
\usepackage{xspace}
\usepackage{mathtools}
\usepackage{wasysym}
\usepackage{marvosym}
\usepackage[table,xcdraw]{xcolor}
\usepackage{lipsum}
\usepackage{authblk}

\usepackage[bookmarks=false]{hyperref}
\cvprfinalcopy % *** Uncomment this line for the final submission

\def\cvprPaperID{3399} % *** Enter the CVPR Paper ID here

\newcommand*\samethanks[1][\value{footnote}]{\footnotemark[#1]}

\title{VirtualHome: Simulating Household Activities via Programs\\[-5mm]}

\author{
Xavier Puig\textsuperscript{1}\thanks{Denotes equal contribution},
~~Kevin Ra\textsuperscript{2}\samethanks[1], 
~~Marko Boben\textsuperscript{3}\samethanks[1], 
~~Jiaman Li\textsuperscript{4}, 
~~Tingwu Wang\textsuperscript{4}, \\ \vspace{-3mm}
~~Sanja Fidler\textsuperscript{4}, 
~~Antonio Torralba\textsuperscript{1} \\
	\textsuperscript{1}MIT \quad
	\textsuperscript{2}McGill University \quad
	\textsuperscript{3}University of Ljubljana \quad
	\textsuperscript{4}University of Toronto  \\
	{\tt\footnotesize \{xavierpuig,torralba\}@csail.mit.edu} \quad
	{\tt\footnotesize kevin.ra@mail.mcgill.ca} \quad 
	{\tt\footnotesize marko.boben@fri.uni-lj.si} \quad
    {\tt\footnotesize \{tingwuwang,ljm,fidler\}@cs.toronto.edu}
}

\begin{document}
\maketitle
\thispagestyle{empty}
\begin{abstract}
In this paper, we are interested in modeling complex activities that occur in a typical household. We propose to use \emph{programs}, i.e., sequences of atomic actions and interactions, as a high level representation of complex tasks. Programs are interesting because they provide a non-ambiguous representation of a task, and allow agents to execute them. However, nowadays, there is no database providing this type of information. Towards this goal, we first crowd-source programs for a variety of activities that happen in people's homes, via a game-like interface used for teaching kids how to code. Using the collected dataset, we show how we can learn to extract programs directly from natural language descriptions or from videos.  We then implement the most common atomic (inter)actions in the Unity3D game engine, and use our programs to ``drive'' an artificial agent to execute tasks in a simulated household environment. Our \emph{VirtualHome} simulator allows us to create a large activity video dataset with rich ground-truth, enabling training and testing of video understanding models. We further showcase examples of our agent performing tasks in our \emph{VirtualHome} based on language descriptions.
\end{abstract}

\vspace{-3mm}
\section{Introduction}
\label{sec:intro}
\vspace{-1mm}

Autonomous agents need to know the sequences of actions that need to be performed in order to achieve certain goals. For example, we might want a robot to clean our room, make the bed, or cook dinner. One can define activities with procedural recipes or programs that describe how one can accomplish the task. A \emph{program} contains a sequence of simple symbolic instructions, each referencing an atomic action (e.g. ``sit'') or interaction (e.g. ``pick-up object'') and a number of objects that the action refers to (e.g., ``pick-up juice''). Assuming that an agent knows how to execute the atomic actions, programs provide an effective means of ``driving'' a robot to perform different, more complex tasks. Programs can also be used as an internal representation of an activity shown in a video or described by a human (or another agent). Our goal in this paper is to automatically generate programs from natural language descriptions, as well as from video demonstrations, potentially allowing naive users to teach their robot a wide variety of novel tasks. 

\begin{figure}[t!]
\vspace{-2.0mm}
\includegraphics[width=\linewidth,trim=0 0 0 0,clip]{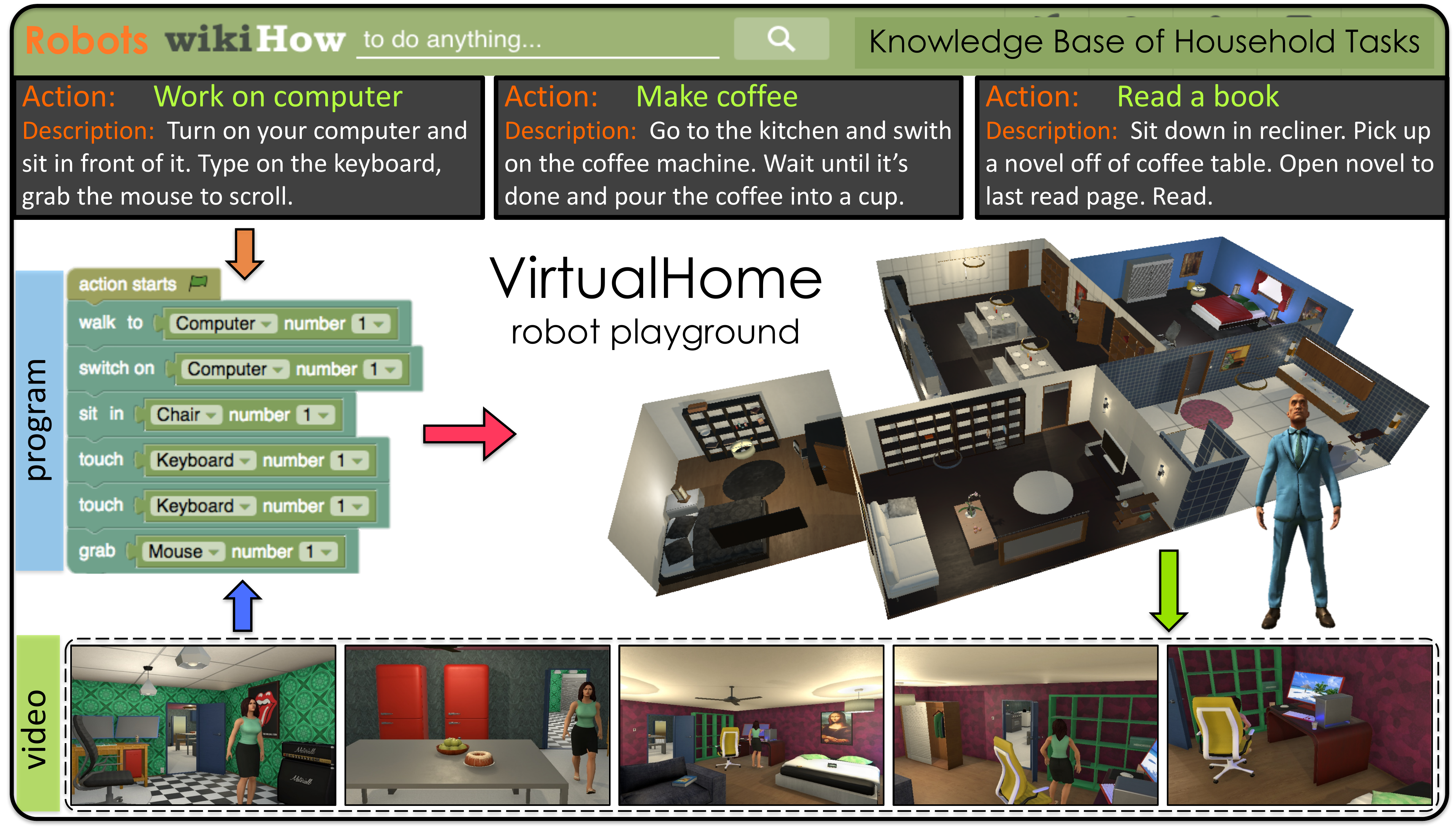}
%\caption{Given a description of an activity in natural language ({\bf top left}), we translate it into a sequence of atomic steps ({\bf bottom left}), which we call a \emph{script}. Our agent executes these scripts in our novel virtual environment that mimics a typical household ({\bf right}).}
\vspace{-7mm}
\caption{\footnotesize We first crowdsource a large knowledge base of household tasks, {\bf (top)}. Each task has a high level name, and a natural language instruction. We then collect ``programs'' for these tasks, {\bf (middle left)}, where the annotators ``translate'' the instruction into simple code. We implement the most frequent (inter)actions in a 3D simulator, called \emph{VirtualHouse}, allowing us to drive an agent to execute tasks defined by programs. We propose methods to generate programs automatically from text {\bf (top)} and video {\bf (bottom)}, thus driving an agent via language and a video demonstration. 
}
%Given \emph{either} a {\bf (1)} a natural language description of an activity ({\bf top left}), or {\bf (2)} a video of a tutor performing an activity ({\bf bottom}), we translate it into symbolic code (seq. of atomic steps) ({\bf middle left}), called a \emph{script}. Our agent executes these scripts in our virtual household environments ({\bf right}).}
\label{fig:intro}
\vspace{-3.0mm}
\end{figure}

Towards this goal, one important missing piece is the lack of a database describing activities composed of multiple steps. We first crowdsource common-sense information about typical activities that happen in people's homes, forming the natural language know-how of how these activities are performed. We then adapt the Scratch~\cite{scratch} interface used for teaching kids how to code in order to collect programs that formalize the activity as described in the knowledge base. %This simple programing interface allowed us to deploy the data collection on Amazon Mechanical Turk and collecting 1440 different activities with programs. 
Note that these programs include \emph{all the steps} required for the robot to accomplish a task, even those that are not mentioned in the language descriptions.  We then implement the most common atomic (inter)actions in the Unity3D game engine, such as \emph{pick-up}, \emph{switch on/off}, \emph{sit}, \emph{stand-up}. By exploiting the physics, navigation and kinematic models in the game engine we enable an artificial agent to execute these programs in a simulated household environment. 

We first introduce our data collection effort and the program based representation of activities. In Sec.~\ref{sec:method} we show how we can learn to automatically translate natural language instructions of activities into programs. In Sec.~\ref{sec:virtualhome} we introduce the \emph{VirtualHome} simulator that allows us to create a large activity video dataset with rich ground-truth by using programs to drive an agent in a synthetic world. Finally, we use the synthetic videos to train a system to translate videos of activities into the program being executed by the agent. Our \emph{VirtualHome} opens an important ``playground'' for both vision and robotics, allowing agents to exploit language and visual demonstration to execute novel activities in a simulated environment. Our data is available online: {\color{magenta}{\href{http://virtual-home.org/}{http://virtual-home.org/}}}.

\begin{figure*}[t!]
\vspace{-3.5mm}
\includegraphics[width=\linewidth,trim=0 238 0 0,clip]{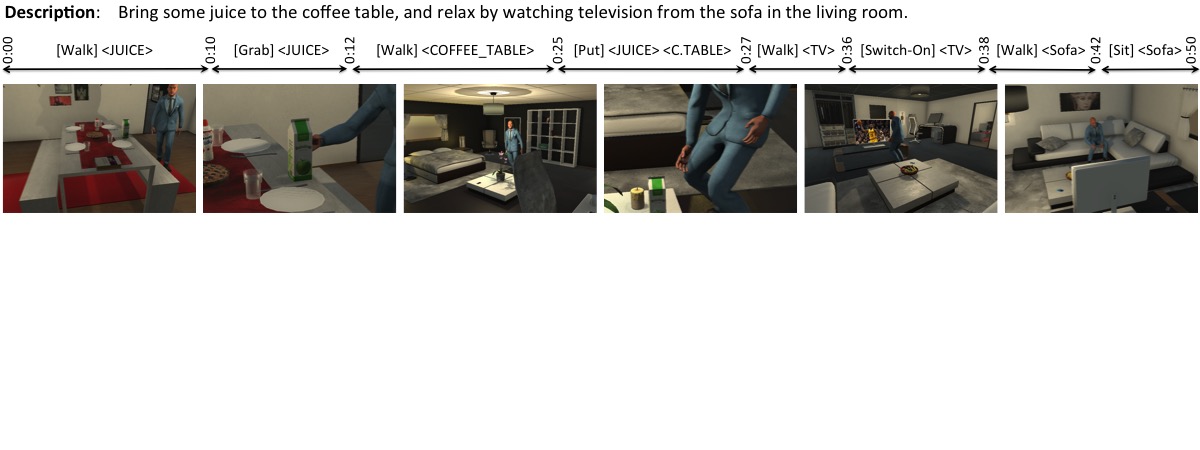}\\
\includegraphics[width=\linewidth,trim=0 310 0 0,clip]{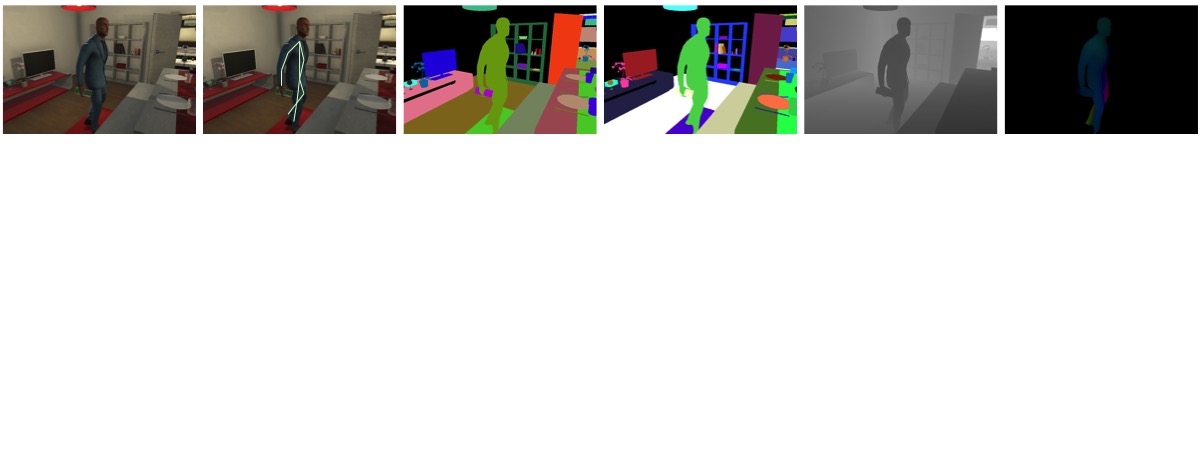}\\[-2mm]
\begin{footnotesize}
\addtolength{\tabcolsep}{25pt}
\begin{tabular}{cccccc}
RGB & pose & class seg. & inst. seg & depth & flow
\end{tabular}
\end{footnotesize}
\vspace{-2mm}
\caption{\small {\bf VirtualHome Activity Dataset}  is a video dataset of composite activities created with our simulator. We start by generating programs using a simple probabilistic grammar. We animate each program in VirtualHome by randomizing the selection of homes, agents, cameras, as well as the placement of a subset of the objects, the initial location of the agent, the speed of the actions, and choice of objects for interactions. Each program  is shown to an annotator who is asked to describe it in natural language {\bf (top row)}. Videos have ground-truth: {\bf (second row)} time-stamp for each atomic action, {\bf (bottom)} 2D and 3D pose, class and object instance segmentation, depth and optical flow.}
\label{fig:dataset}
\vspace{-3.0mm}
\end{figure*}

\vspace{-2.0mm}
\section{Related Work}
\label{sec:related}
\vspace{-1mm}

{\bf Actions as programs.} A few works have defined activities as programs. In~\cite{Yang15}, the authors detect objects and actions in cooking videos and generate an ``action plan'' using a probabilistic grammar. By generating the plan, the robots  were able to execute complex actions by simply watching videos. These authors further collected a tree bank of action plans from annotated cooking videos~\cite{Yang14}, creating a knowledge base of actions as programs for cooking. \cite{Nyga12} tried to translate cooking recipes into action plans using an MRF. \cite{saxena15,Alayrac16} also argued for actions as a sequence of atomic steps. They aligned YouTube how-to videos with their narrations in order to parse videos into such programs. Most of these works were limited to either a small set of activities, or to a narrow domain (cooking). We go beyond this by creating a knowledge base about an exhaustive set of activities and tasks that people do in their homes. 

\iffalse
{\bf Video parsing}. Recent work has demonstrated that alignment of long video sequences and related text is possible with reasonable success~\cite{ZhuICCV15,Tapaswi_J1_PlotRetrieval,Tapaswi2015_Book2Movie}. \cite{saxena15,Alayrac16} aligned YouTube how-to videos with their narrations in order to parse videos into a sequence of actions. This work is related to ours in that it tries to break down complex actions into smaller atomic units. However, it assumes that videos are narrated which is usually not the case for the common activities humans perform every day. 
In~\cite{Yang15}, the authors detect objects and actions in cooking videos and generate an ``action plan'' using a probabilistic grammar. This is an exciting first attempt at having robots learn to execute complex actions by watching videos. The authors further collected a tree bank of action plans from annotated cooking videos~\cite{Yang14}. Here, we crowd-source the scripts and learn to generate them from descriptions and simulated videos, giving us access to ``free'' ground-truth on a larger scale. Related is also the effort in~\cite{Nyga12}, which tries to translate cooking recipes into action plans using an MRF.
\fi

~\cite{Sigurdsson16} crowd-sourced scripts of people's actions at home in the form of natural language. These were mostly comprised of one or two sentences describing a short sequence of actions. While this is valuable information, language is very versatile and thus hard to convert into a usable program on a robot. We show how to do this in our work. 
%, and Turkers were asked to act out the script in front of a camera. %This interesting data collection yielded 10K videos of simple indoor activities described with natural language. 
%Here we take a different approach aiming to simulate a number of such composite activities. 

{\bf Code generation.} There is increased interest in generating and interpreting source code~\cite{Li17}. Work most relevant to ours produces code given natural language inputs.~\cite{Allamanis15} retrieves code snippets from \emph{Stackoverflow} based on language queries. Given a sentence describing conditions,~\cite{Quirk15} produces If-This-Then-That code.~\cite{Ling16} generates a program specifying the logic of a card game given a short description of the rules. In~\cite{Johnson17}, the authors inferred programs to answer visual questions about images.  Our work differs in the domain, and works with text or video as input. 
%\textcolor{red}{Talk about clevr? Generate programs for VQA}
%the authors propose an attention model on an structured input representing various properties of a game character, and generate short snippets of code 

{\bf Robotics}. A subfield of robotics aims at teaching robots to follow instructions provided in natural language by a human tutor.  However, most of the existing literature deals with a constrained problem, for example, they learn to translate navigational instructions into a sequence of robotic actions~\cite{Tellex11,MacMahon06,Lauria01,mei2016navigational}. These instructions are typically simpler as they directly mention what to do next, and the action space is small. This is not the case in our work which also considers interactions with objects, and everyday activities which are typically far more complex. %The approach taken in~\cite{mei2016navigational} is related to our script decoder in that it exploits an encoder-decoder neural architecture.  Our approach incorporates action-object relations and language priors into the network which we show to significantly outperform their baseline. 

{\bf Simulation}. Simulations using game engines have been developed to facilitate training visual models for autonomous driving~\cite{Gaidon:Virtual:CVPR2016,Richter16,Dosovitskiy17}, quadcopter flying~\cite{AirSim}, or other robotic tasks~\cite{openaigym}. Recent works have focused on simulating indoor environments, allowing for target-driven indoor navigation or interactive question answering~\cite{ai2thor, house3d, HOME, savva2017minos}. A few of these works~\cite{ai2thor, HOME} include actionable objects, allowing to interact and change the environment. Our work focuses on simulating a wide range of human actions, both in terms of objects interactions and human poses, which allows to simulate common activities. We are not aware of simulators at the scale of objects and actions in a home, like ours. Lastly, we give credit to the popular game Sims which we draw our inspiration from. Sims is a strategic video game mimicking daily household activities. Unfortunately, the source of the game is not public and thus cannot be used for our purpose.

%Simulation-based datasets for object manipulation have been collected in robotics~\cite{Abbeel15}. Simulations using game engines have recently been developed to facilitate training visual models for autonomous driving~\cite{Gaidon:Virtual:CVPR2016,Richter16} and quadcopter flying~\cite{AirSim}. Several simulators have also been released for the purpose of reinforcement learning, such as MuJoCo and OpenAI's Gym. These feature various 2D games, or 3D human dynamics. Recently,~\cite{zhu16} released a simulator for target-driven indoor navigation, which however does not simulate activities. %The latter aims  at teaching agents how to move as well as grab very simple objects. 
%We are not aware of simulators at the scale of objects and actions in a home. Lastly, we give credit to the popular game Sims which we draw our inspiration from. Sims is a strategic video game mimicking daily household activities. Unfortunately, the source of the game is not public and thus cannot be used for our purpose.

%However, for operation in the real world, a robot needs to also be able to interact with objects and humans in complex ways. This is particularly important for household robotics  
%where the robot needs to pick up and utilize many objects for a variety of tasks.

%\section{Dataset}

\vspace{-2mm}
%\section{Action scripts: a database of common actions}
\section{KB of Household Activities for Robots}
\label{sec:tasks}
\vspace{-1mm}

Our goal is to build a large repository of common activities and tasks that we perform in our households in our daily lives. These  tasks can include simple actions like ``turning on TV" or complex ones such as ``make coffee with milk". What makes our effort unique is that we are interested in collecting this information for robots. Unlike humans, robots need more direct instructions. For example, in order to ``watch tv'', one might describe it (to a human) as ``Switch on the television, and watch it from the sofa''. Here, the actions ``grab remote control'' and ``sit/lie on sofa'' have been omitted, since they are part of the commonsense knowledge that humans have. In our work, we aim to collect {\bf all the steps} required for a robot to successfully execute a task, including the commonsense steps. In particular, we want to collect programs that fully describe activities. %We additionally want to collect natural language instructions, and learn how to translate these into programs. 

%all the steps that are required to execute a task. Most of these steps are common sense knowledge for humans, and would not even be mentioned if one was to instruct another human. For example, in order to ``watch tv'', one might describe it as ``Switch on the television, and watch it from the sofa''. Here, the action sit/lie on the sofa has been omitted, since it's part of the common sense knowledge that we have. Our goal here is to collect exhaustive information about each activity including common sense steps. 

Describing actions as programs has the advantage that it provides a clear and non-ambiguous description of all the steps needed to complete a task. Such programs can then be used to instruct a robot or a virtual character. Programs can also be used as a representation of a complex task that involves a number of simpler actions, providing a way to understand and compare activities and goals.  

\subsection{Data Collection}

\begin{figure*}
\begin{center} 
\begin{minipage}{0.55\linewidth}
   \includegraphics[width=1\linewidth]{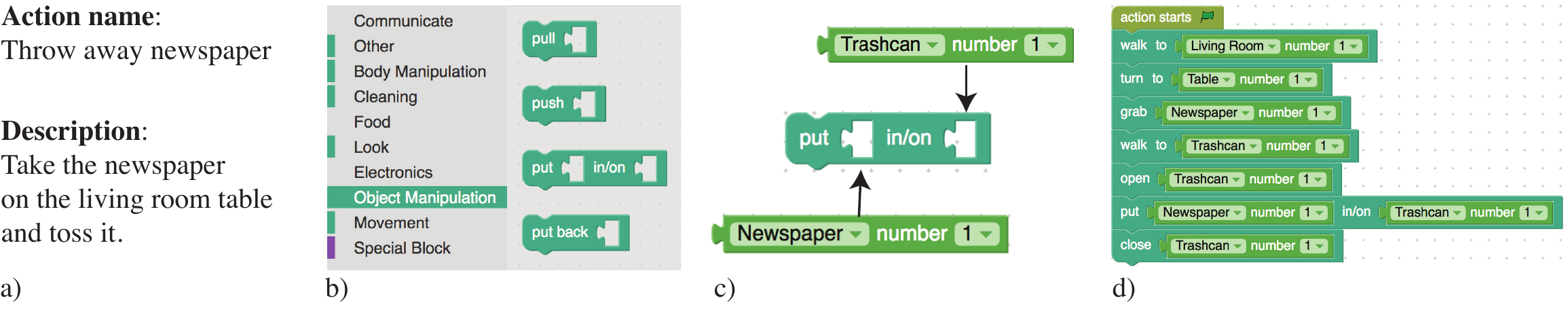}
   \vspace{-6mm}
      \caption{\small a) Description provided by a worker. b) User interface showing the list of block categories and 4 example blocks, c) Example of composition of a block by adding the arguments. Each block is like a Lego piece where the user can drop arguments inside and attach one block to another. d) Final program corresponding to the description from (a).}
\label{fig:interfazeScripts} 
   \end{minipage}
   \hspace{2mm}
   \begin{minipage}{0.425\linewidth}
   \vspace{-2mm}
    \includegraphics[width=1\linewidth,trim=0 0 160 0,clip=true]{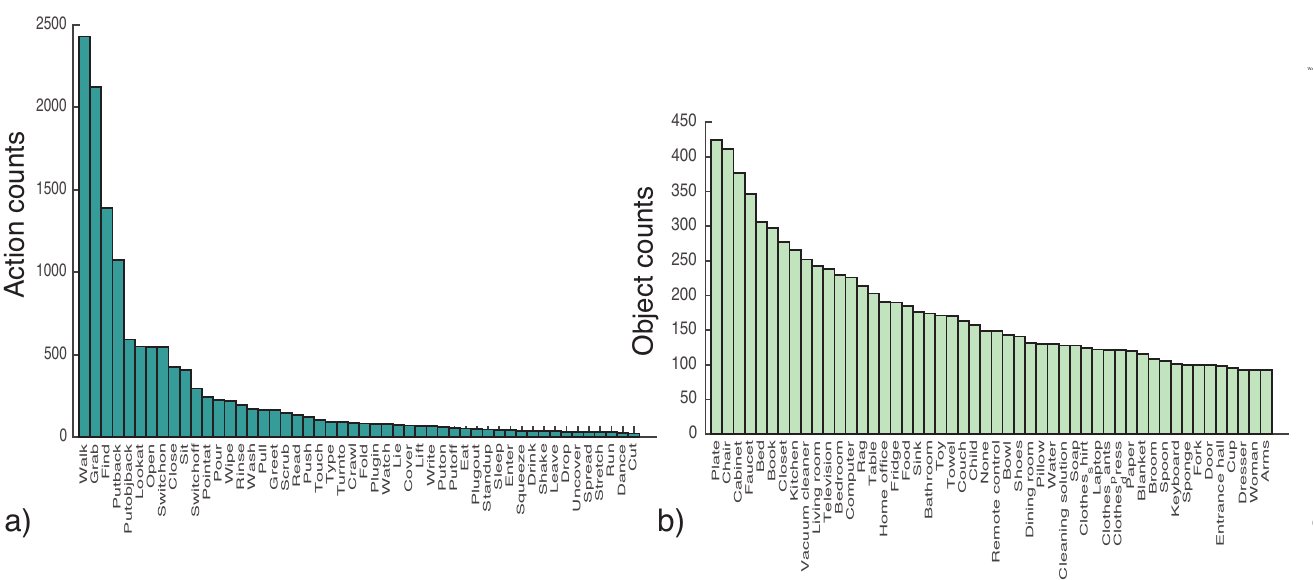}
    \vspace{-8mm}
       \caption{\small  a) Counts of actions in our \emph{ActivityPrograms} dataset, b) object counts (zoom to read)}
       %, c) similarity matrix of activities based on the similarity between programs (zoom to read).}
\label{fig:stats} 
     \end{minipage}
\end{center} 
\vspace{-8mm}
\end{figure*}

In this section, we describe our dataset collection using crowdsourcing.  
Describing actions as programs can be a challenging task as most annotators have no programing experience. We split the data collection effort in two parts. In the first part, we ask AMT workers to provide verbal descriptions of daily household activities. In particular, each worker is asked to come up with a common activity/task, give it a high level name, eg ``make coffee'', and describe it in detail.  
In order to cover a wide spectrum of activities we pre-specified in which scene the activity should start. Scenes were selected randomly from a list of $8$ scenes (\emph{living room}, \emph{kitchen}, \emph{dining room}, \emph{bedroom}, \emph{kids bedroom}, \emph{bathroom}, \emph{entrance hall}, and \emph{home office}). %An example is: activity name, {\it Read an email}; description, {\it Turn on computer. Wait for it to load. Get online. Go to the email service. Open the email. Read the email.} The issue with verbal descriptions is that they might be ambiguous or miss important steps that are common sense for humans. 
An example of a described activity is shown in Fig.~\ref{fig:interfazeScripts}. Note that these descriptions may likely omit the commonsense steps, as they were written by ``naive'' workers that were describing these activities as they would to a (human) friend.

In the second stage, we showed the collected descriptions to the AMT workers and asked them to translate these descriptions into programs using a graphical programing language. We told them to produce a program that will ``drive'' a robot to successfully accomplish the described activity. Our interface builds on top of MIT's Scratch project~\cite{scratch} designed to teach young children to write symbolic code. We found that workers were capable of quickly learning to produce useful programs  by  providing them with a carefully designed tutorial. Fig.~\ref{fig:interfazeScripts} shows a snapshot of the programing interface. Finally, we asked more qualified workers hired via Upwork crowdsourcing platform to double check the collected data.

Workers had to compose a program by composing a sequence of steps. Each instruction is a Scratch block from a predefined list of 77 possible blocks compiled by analyzing the frequency of verbs in the collected descriptions. Each step in the program is defined by a block. A block defines a syntactic frame with an action and a list of arguments (e.g., the block \emph{walk} requires one argument to specify the destination, Fig.~\ref{fig:interfazeScripts}.c). To simplify the search for blocks they are organized according to 9 broad action categories (Fig.~\ref{fig:interfazeScripts}.b). 

We required that the program contains all the steps, even those not explicitly mentioned in the description, but that could be inferred from common-sense. Fig.~\ref{fig:interfazeScripts}.d shows an example of a  program.  We also allowed annotators to use a ``special'' block for missing actions, where the step can be written as free-form text. Programs using this special block will not be used in the rest of the paper, but allowed us in identifying new blocks that needed to be added. 

More precisely, step $t$ in the program can be written as:
\vspace{-0.2mm}
\begin{equation}
\mathrm{step}_t = [action_{t}]\ \ \langle object_{t,1} \rangle (id_{t,1}) \ ... \ \langle object_{t,n}\rangle (id_{t,n})\nonumber
\label{eq:step}
\end{equation}
Here, $id$ is an unique identifier (counter) of an object and helps in disambiguating different instances of objects that belong to the same class.  An example of a program for ``watch tv'' would be:
\vspace{-1mm}
\begin{center}
\begin{small}
\begin{tabular}{l}
$\mathrm{step}_1 = [\mathrm{Walk}] \ \ \langle\mathrm{TELEVISION}\rangle (1)$\\
$\mathrm{step}_2 = [\mathrm{SwitchOn}] \ \ \langle\mathrm{TELEVISION}\rangle (1)$\\
$\mathrm{step}_3 = [\mathrm{Walk}] \ \ \langle\mathrm{SOFA}\rangle (1)$\\
$\mathrm{step}_4 = [\mathrm{Sit}] \ \ \langle\mathrm{SOFA}\rangle (1)$\\
$\mathrm{step}_5 = [\mathrm{Watch}] \ \ \langle\mathrm{TELEVISION}\rangle (1)$\\
\end{tabular}
\end{small}
\vspace{-1mm}
\end{center}
Here, the programs defines that the television in steps 1, 2 and 5 refer to the same object instance. 

%\begin{figure}[t!]
%\vspace{-4mm}
%\includegraphics[width=0.7\linewidth,trim=380 90 120 80,clip] {figures/affordance_graph.eps} 
%\vspace{-6mm}
%\caption{\small Co-occurr. of actions-objects in \emph{ActivityScripts} dataset.}
%\label{fig:affordance}
%\vspace{-4mm}
%\end{figure}

\vspace{-0mm}
\subsection{Dataset Analysis}
\label{sec:dataset}

In the first part we collected 1814 descriptions. From those, we were able to  collect programs for 1703 descriptions.  Some of the programs contained several ``special blocks'' for missing actions, which we remove, resulting in 1257 programs. We finally selected a set of tasks and asked workers to write programs for them, obtaining 1564 additional programs. The resulting 2821 programs form our \emph{ActivityPrograms} dataset. On average, the collected descriptions have 3.2 sentences and 21.9 words, and the resulting programs have 11.6 steps on average. The dataset statistics are summarized in Table~\ref{tab:stats}.a.

The dataset covers 75 atomic actions and 308 objects, making 2709 unique steps. Fig.~\ref{fig:stats}.a shows a histogram of the 50 most common actions appearing in the dataset, and, Fig.~\ref{fig:stats}.b, the 50 most common objects.  %Fig.~\ref{fig:affordance} shows co-occurrence of actions-objects in instructions (steps in programs), with the intensity of the edges representing how common a given action-object pair is.

Our dataset contains activities with several examples, and we analyze their diversity by comparing their programs. Table~\ref{tab:stats}.b analyzes 4 selected activities. We compute their similarities as the average length of the longest common subsequences computed between all pairs of programs. 

We can also measure distances between activities by measuring the distance between programs. The similarity between two programs is measured as the length of their longest common subsequence of instructions divided by the length of the longest program. Table~\ref{tab:stats}.c. shows the similarity matrix (sorted to better show the block diagonal structure) between different activities in our dataset.  

{\bf Completeness of programs.} We analyze whether the collected programs contain all the necessary steps to execute the given task. We sample 100 collected programs and ask 5 AMT workers to rate whether the program is complete, missing minor steps (sitting in a chair before walking towards it) or important steps (filling a glass before drinking). Results show that 64\% of the programs are complete, 28\% are missing minor steps and 8\% are missing crucial steps. %We can correct many of the minor steps by adding walking actions before interacting with objects.

% We then animated each script in our simulator, and automatically generated ground-truth which allows us to train and evaluate our models. We split the data into trainval and test. %We use slightly different splits for text and video-based tasks: for video-based tasks we make a larger test allowing us to evaluate performance on two subsets of test, one with homes seen in training and one with unseen homes. Note that this makes the accuracies for text-based and video-based script prediction related, but not directly comparable. 
%As can be seen from Table~\ref{tab:stats}, the descriptions for the \emph{VirtualHome Activity} dataset are of comparable length. However, 
%the vocabulary here was biased towards the vocabulary used in scripts. This also shows later in results. 

\begin{table*}[t!]
\vspace{-3mm}
\begin{minipage}{0.38\linewidth}
\vspace{2mm}
\begin{footnotesize}
\addtolength{\tabcolsep}{-4.9pt}
\begin{tabular}{|l|c|c|c|c|}
\hline
Dataset & \# prog. & avg \# steps & avg \# sent. & avg \# words\\
\hline
ActivityProg.&  2821 &  11.6 & 3.2 & 21.9 \\
SyntheticProg. &  5193 & 9.6 & 3.4 & 20.5 \\
\hline
\end{tabular}
\end{footnotesize}
\vspace{-1mm}
\begin{center}
{\footnotesize (a)}
\end{center}
\end{minipage}
\hspace{3mm}
\begin{minipage}{0.32\linewidth}
\vspace{1mm}
\begin{footnotesize}
\addtolength{\tabcolsep}{-3pt}
\begin{tabular}{|cccc|}
\hline
Action & \# Prog. & LCS & Norm. LCS \\
\hline
Make coffee & 69 & 4.56 & 0.26\\
Fold laundry & 11 & 1.29 & 0.08 \\
Watch TV & 128 & 3.65 & 0.40\\
Clean & 42 & 0.76 & 0.04 \\
\hline
\end{tabular}
\end{footnotesize}
\begin{center}
{\footnotesize (b)}
\end{center}
\end{minipage}
{\footnotesize (c)}\begin{minipage}{0.175\linewidth}
%\includegraphics[width=1\linewidth,trim=380 40 0 0,clip=true]{newfigures/data_stats_reduced.pdf}
%\scalebox{1}{\includegraphics[width=1\linewidth,trim=120 180 200 350, clip=true]{newfigures/affinity_reduced_new.pdf}}
\scalebox{1.4}{\includegraphics[width=1\linewidth,trim=50 140 200 340, clip=true]{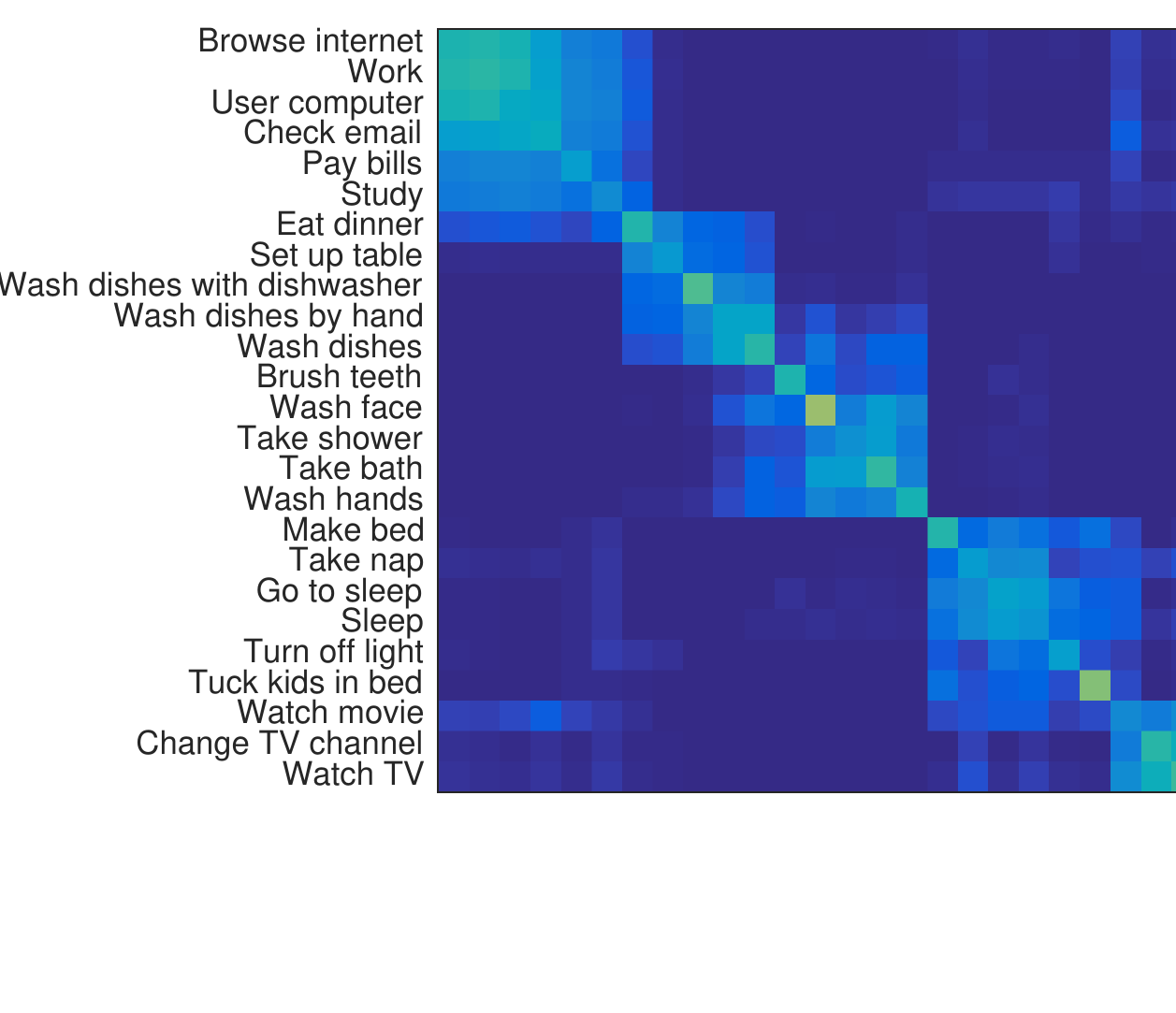}}
\end{minipage}
\vspace{-10mm}
\caption{\small {\bf (a)} We analyze programs and natural language descriptions for both, real activities in \emph{ActivityPrograms} (Sec.~\ref{sec:tasks}), and synthesized programs (with real descr.). {\bf (b)} \emph{ActivityPrograms}: Analyzing diversity in the same activity, by computing similarities across all pairs of the collected programs. ``LCS" denotes longest common subsequence. For ``norm.LCS", we normalize by max length of the two programs.
 {\bf (c)} shows the similarity matrix (sorted to better show the block diagonal structure) between different activities in our dataset.  
}
\label{tab:stats}
\vspace{-3mm}
\end{table*}

\vspace{-2mm}
\section{\emph{VirtualHome}: Simulator of Household Tasks}
\label{sec:virtualhome}
\vspace{-1mm}

The main motivation behind using programs to represent activities is to ``drive'' robots to perform tasks by having them executing these programs. As a proxy, we here use programs to 
%One interesting application of having programs describing activities is that we can use them to 
drive characters in a simulated 3D environment. Simulations are useful as they define a playground for ``robots'', an environment where artificial agents can be taught to perform tasks. % using visual information. This is a very challenging problem on its own which we plan to tackle in the future. 
Here, we focus on building the simulator, and leave learning inside the simulator to future work. In particular, we will assume the agent has access to all 3D and semantic information about the environment, as well as to manually defined animations. Our focus will be to show that programs represent a good way of instructing such agents. 
%using it to showcase that programs can drive robots to perform tasks when having access to the full 3D and semantic information about the environment. Furthermore, 
%allow robots to train themselves in performing complex tasks using 
Furthermore, our simulator will allow us to generate a large-scale video dataset of complex activities that is rich and diverse. We can create such a dataset by simply recording the agent executing programs in the simulator. The simulator then provides us with dense ground-truth information, eg semantic segmentation, depth, pose, etc. Fig.~\ref{fig:dataset} showcases this dataset.

%Such a video dataset allows for training and testing complex video understanding models on a larger scale, and help us understand their performance better in a more controlled setting. %Recent work has explored simulated data for autonomous driving~\cite{Gaidon:Virtual:CVPR2016,Richter16}. 

\begin{figure*}[t!]
\vspace{1mm}
\begin{minipage}{0.665\linewidth}
\addtolength{\tabcolsep}{-3.7pt}
\begin{tabular}{cccccc}
\iffalse
\includegraphics[height=0.188\linewidth,trim=5 65 20 50,clip]{figs/apts/scene0rot.png}  &\includegraphics[height=0.19\linewidth,trim=20 95 40 100,clip]{figs/apts/scene4rot.png} &
\includegraphics[height=0.188\linewidth,trim=30 5 30 5,clip]{figs/apts/scene2rot.png}  & \includegraphics[height=0.19\linewidth,trim=70 5 40 5,clip]{figs/apts/scene5rot.png} & \includegraphics[height=0.188\linewidth,trim=30 5 0 5,clip]{figs/apts/scene6rot.png} & \includegraphics[height=0.19\linewidth,trim=0 5 0 5,clip]{figs/apts/scene3rot.png}% &\includegraphics[height=0.19\linewidth,trim=20 20 20 0,clip]{figs/apts/chars.png} \\
\fi
\includegraphics[height=0.23\linewidth,trim=5 65 20 50,clip]{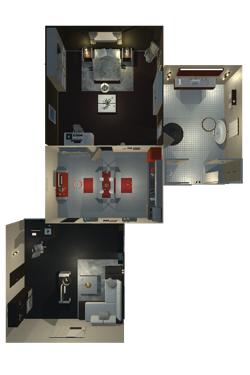}  &\includegraphics[height=0.23\linewidth,trim=20 95 40 100,clip]{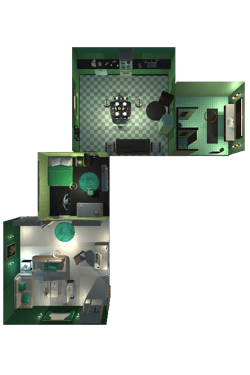} &
\includegraphics[height=0.23\linewidth,trim=30 5 30 5,clip]{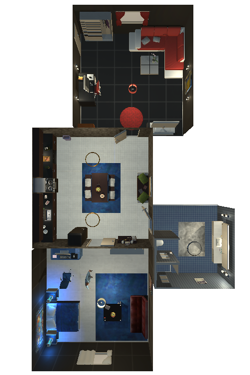}  & \includegraphics[height=0.23\linewidth,trim=70 5 40 5,clip]{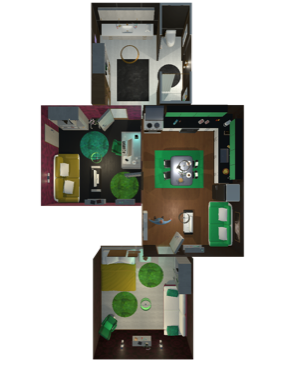} & \includegraphics[height=0.23\linewidth,trim=30 5 0 5,clip]{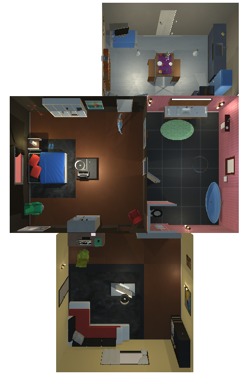} & \includegraphics[height=0.23\linewidth,trim=0 5 0 5,clip]{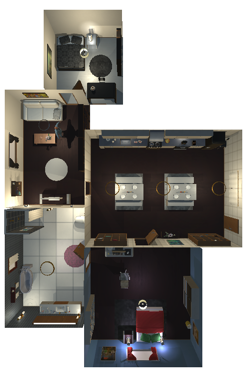}
\end{tabular}
\vspace{-2.5mm}
\caption{\small 3D households in our VirtualHome. Notice the diversity in room and object layout and appearance. Each home has on average $357$ objects. First $4$ scenes are used for training, the fifth is also used in val, and all scenes are used when testing our video-to-script model.}
\label{fig:scenes}
\vspace{-0mm}
\end{minipage}\hspace{4mm}
\begin{minipage}{0.304\linewidth}
\addtolength{\tabcolsep}{-1.3pt}
\begin{center}
\begin{tabular}{cccc}
\includegraphics[height=0.415\linewidth]{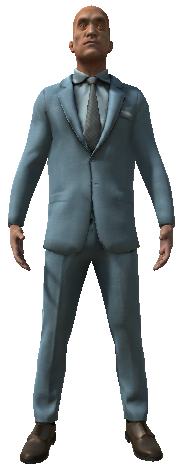} & \includegraphics[height=0.41\linewidth,trim=0 0 0 00,clip=true]{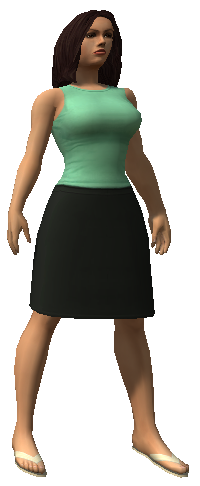} &\includegraphics[height=0.415\linewidth,trim=10 0 0 00,clip=true]{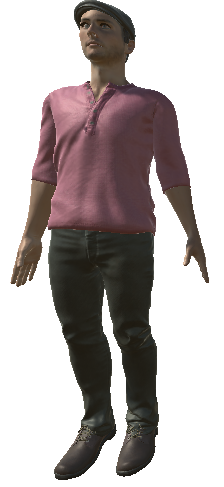} &
\includegraphics[height=0.415\linewidth]{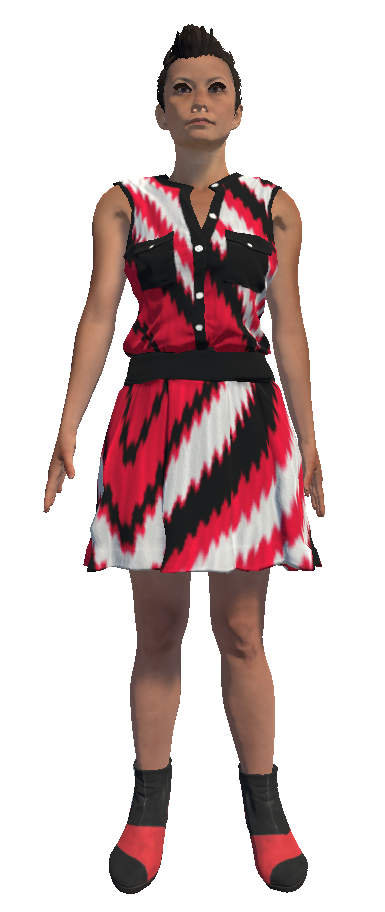}\\[-2mm]
\scriptsize{male 1} & \scriptsize{female 1} & \scriptsize{male 2} & \scriptsize{female 2}
\end{tabular}
\end{center}
\vspace{-4.5mm}
\caption{\small Agents in VirtualHome. We use \emph{male 1} and \emph{female 1} in train., and all agents when testing our video-to-program model.}
\label{fig:agents}
\end{minipage}
\vspace{-2mm}
\end{figure*}

We implemented our \emph{VirtualHome} simulator using the Unity3D game engine which allows us to exploit its kinematic, physics and navigation models, as well as 
user-contributed 3D models available through Unity's Assets store. We obtained six furnished homes and 4 rigged humanoid models from the web. On average, each home contains $357$ object instances ($86$ per room).  We collected objects from additional $30$ object classes that appear in our collected programs yet are not available in the package, via the 3D warehouse~\footnote{https://3dwarehouse.sketchup.com}. To ensure visual diversity, we collected at least 3 different models per class. The apartments and agents are shown in Fig.~\ref{fig:scenes} and Fig.~\ref{fig:agents}. 

\subsection{Animating Programs in VirtualHome}

Every step in the program requires us to animate the corresponding (inter)action in our virtual environment. We thus need to both, determine which object in the home (which we refer to as the \emph{game object}) the step requires as well as properly animating the action. To get the former we need to solve an optimization problem by taking into account all steps in the program and finding a feasible path. For example, if the program requires the agent to switch on a computer and type on a keyboard, ideally the agent would type on the keyboard next to the chosen computer and not navigate to another keyboard attached to a different computer in possibly a different room. %We describe here all the steps needed to animate characters. 
We now describe our simulator in more detail.

%We discuss our optimization in Sec.~\ref{sec:execution}. For the latter, we assume we know which game object we need to interact with and the task is to animate the interaction. The full action then includes walking to a room or object, and interacting with it in different ways. We discuss this next.

%We use Unity's standard assets for the walking and running character animation, and the built-in NavMesh framework for navigation. The latter plans a path from point A to point B by avoiding any obstacles along the path, and animates the walk/run action in a physically plausible way. 

{\bf Animating atomic actions.} There is a huge variety and number of atomic actions that appear in the collected programs, as can be seen in Fig.~\ref{fig:stats}. We implemented the $12$ most frequent ones: \emph{walk/run}, \emph{grab}, \emph{switch-on/off}, \emph{open/close}, \emph{place}, \emph{look-at}, \emph{sit/standup}, \emph{touch}. 
Note that there is a large variability in how an action is performed depending on to which object it is applied to (e.g., opening a fridge is different than opening a drawer). 
We use Unity's NavMesh framework for navigation (path planner to avoid obstacles). For each action we compute the agent's target pose and animate the action using RootMotion FinalIK inverse kinematics package.  We further animate certain objects the agent interacts with, e.g.,  we shake a coffee maker, animate toast in a toaster, show a (random) photo on a computer or TV screen, light up a burner on a stove, and light up the lamps in the room, when these objects are switched on by the agent.

{\bf Preparing the Scene.} While every 3D home already contains many objects, the programs may still mention objects that are not present in the scene. To deal with this, we first ``set'' the scene by placing all missing objects that a program refers to in the home, before we try to execute the program.  
To be able to prepare a scene in a plausible way, we collect a knowledge base of possible object locations. The annotator is shown the class name and selects a list of other objects (including \emph{floor}, \emph{wall}) that are likely to support it. 
%We allow the annotator to also add room information, e.g., \emph{floor: kids bedroom}.

%Given a program, we place all missing objects by randomly selecting 1) the 3D model for the object class, 2), target surface for each missing object by exploiting our knowledge base. %We exploit the objects' 
%(the new model as well as for those in the scene) 
%colliders, which are essentially 3D bounding boxes around the object. 
%We compute free space on the target surface, and randomly place the new object in the available space. 

{\bf Executing a Program.} To animate a program we need first to create a mapping between the objects in the program and the corresponding instances inside the virtual simulator. Furthermore, for each step in the program, we also need to compute the interaction position of the agent with respect to an object, and any additional information needed to animate the action (e.g., which hand to use, speed).
We build a tree of all possibilities of assigning game objects to objects in the program, along with all interaction positions and attributes. To traverse the tree of possible states we use backtracking and stop as soon 
as a state executing the last step is found. Since the number of 
possible object mappings for each step is small, and we can prune the number 
of interaction positions to a few, our optimization runs in a few seconds, on average.

{\bf Animation.} We place 6-9 static cameras in each room, $26$ per home on average. During recording, we switch between cameras based on agent's visibility. In particular, we randomly select a camera which sees the agent, and keep it until the agent is visible and within allowed distance. For agent-object interaction we also try to select a camera and adjust its field of view to enhance the visibility of the interaction. We further randomize the position, angle and field of view of each camera. Randomization is important when creating a dataset to ensure diversity of the final video data. 

%{\bf VirtualHome Activity Dataset.} Due to large variability in activities we collected, many programs cannot be animated (missing atomic actions or objects). We thus automatically synthesize programs restricted to our implemented atomic actions, using a simple probabilistic grammar. %Details are given in Suppl. material. 
{\bf VirtualHome Activity dataset}. Since the programs in \emph{ActivityPrograms} represent real activities that happen in households, they contain significant variability in actions and objects that appear in steps. While our ultimate aim is to be able to animate all these actions in our simulator, our current efforts only support the top 12 most frequent actions. We thus create another dataset that contains programs containing only these actions in their steps. The creation of this dataset is explained below.

We synthesized 5,193  programs using a simple probabilistic grammar, and had each one described in natural language by a human annotator. Although these programs were not given by annotators, they produced reasonable activities, creating a much larger dataset of paired descriptions-programs at a fraction of the cost. We then animated each program in our simulator, and automatically generated ground-truth which allows us to train and evaluate our video models.
As can be seen from Table~\ref{tab:stats}, descriptions in \emph{VirtualHome Activity} dataset are of comparable length. However, 
the vocabulary here was biased towards that used in programs.

We animate the programs as described above, by randomizing the selection of home, an agent, cameras, placement of a subset of objects, initial location of the agent, speed of the actions, and choice of objects for interactions. We build on top of~\cite{unitylink} to automatically generate groundtruth: 1) time-stamp of each step to video, 2) agent's 2D/3D pose, 3) class and instance segmentation, 4) depth, 5) optical flow,  6) camera parameters. Example of data is shown in Fig.~\ref{fig:dataset}.

\vspace{-1.5mm}
\section{From Videos and Descriptions to Programs}%Text and Videos}
\label{sec:method}
\vspace{-0.5mm}

\begin{figure*}[t!]
\vspace{-3mm}
\centering
\begin{minipage}{0.8\linewidth}
\includegraphics[width=0.97\linewidth,trim=10 152 0 48,clip]{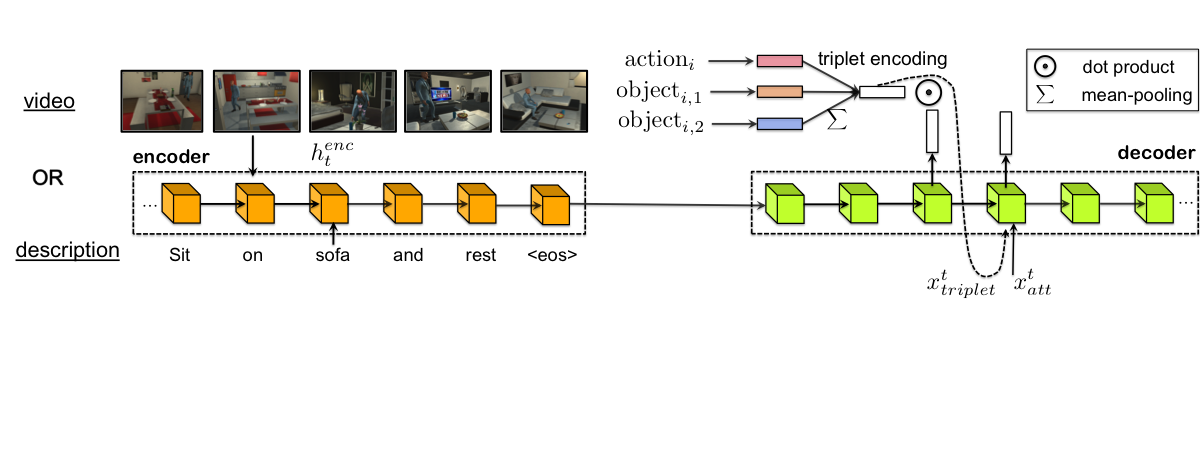}
%\caption{Our encoder-decoder RNN model for generating scripts from textual descriptions and video. }
%\label{fig:model}
\end{minipage}\hspace{0.2mm}
%\begin{minipage}{0.273\linewidth}
%\includegraphics[width=0.97\linewidth,trim=20 105 630 10,clip]{figs/model/Slide9}
%\includegraphics[width=1.\linewidth,trim=0 152 430 50,clip]{figs/model/video_model}
%\caption{Our encoder-decoder RNN model for generating scripts from textual descriptions and video. }
%\caption{We partition the video into 2-sec long clips and encode each clip with an LSTM. The input at each time step is a segmentation map (the softmax output from a DilatedNet applied to the frame), and an attention vector. See text for details.}
%\label{fig:modelVideo}
%\end{minipage}
\vspace{-2mm}
\caption{\small Our encoder-decoder LSTM for generating programs from natural language descriptions or videos.}
\label{fig:modelVideo}
%\vspace{-2mm}
\end{figure*}

We introduce a novel task using our dataset. In particular, we aim to generate a program for the activity from either a natural language description or from a video demonstration. 
%In particular, our goal is to generate a program for an activity from a video, i.e., inferring a sequence of atomic steps by observing another agent performing an activity. Furthermore, we also aim to generate a script when given a description of an activity written in natural language. Note that once we have a script, a new agent could potentially execute the activity in a possibly novel environment.
%Furthermore, we also want to be able to generate scripts from videos, i.e., inferring a sequence of executable steps by observing another agent performing an action. 
We treat the task of transcribing an input (description or video) 
into a program as a translation problem. 
We adapt the seq2seq model~\cite{sutskever2014sequence} for our task, and train it with Reinforcement Learning that exploits the reward from the simulator. %In particular, our model consists of an RNN encoder that encodes the input sequence into a hidden vector representation, and another RNN acting as a decoder, generating one step of the program  at a time. We use an LSTM with 100-dim hidden states as our encoder. %We use a one-hot encoding of each word in a description, project it with a linear layer and feed it into the LSTM.

Our model consists of an RNN encoder that encodes the input sequence into a hidden vector representation, and another RNN acting as a decoder, generating one step of the program at a time. We use LSTM with 100-dim hidden states as our encoder.  At each step $t$, our RNN decoder decodes a step which takes the form of eq.~\eqref{eq:step}. Let ${\bf x}_t$ denote an input vector to our RNN decoder at step $t$, and let $h^t$ be the hidden state. Here, $h^t$ is computed as in the standard LSTM using $\tanh$ as the non-linearity. Let $a_i$ be a one-hot encoding of an action $i$, and $o_i$ a one-hot encoding of an object. We compute the probability $p_i^t$ of an instruction $i$ at step $t$ as:
\vspace{-1.0mm}
\begin{align}
\label{eq:triplet}
\tilde a_i=W_a a_i,\:\:\tilde o_{i,n}&=W_o o_{i,n},\:\:
v_i =\mathrm{mean}(\tilde a_i,\tilde o_{i,1},... , \tilde o_{i,n}) \nonumber \\
p_i^t&=\mathrm{softmax}_i(\frac{v_{i}}{\|v_{i}\|}^T\cdot W_{v}(h^t \| {\bf x}_{t}^{att}))
\end{align}
where $W_a$ and $W_o$ and $W_{v}$ are learnable matrices, and $v_i$ denotes an embedding of an instruction. Note that here, $n$ is either $1$ or $2$ (our instructions have at most two objects). 
%Notice that $v_i^T\cdot h^t$ decomposes into individual terms, i.e., $\tilde a_i^T\cdot h^t$, $\tilde o_{i,1}^T\cdot h^t$, and $\tilde o_{i,2}^T\cdot  h^t$, and we can thus compute it efficiently for all triplets, by first pre-computing the dot products for all actions and objects, and then score each triplet by looking up the corresponding dot products, and summing. Thus, even though the number of triplets can be very large, our computation is efficient. Since some actions  require fewer objects than two, we use a token \emph{none} to encode the ``missing'' objects in a triplet.

The input vector ${\bf x}_t$ concatenates multiple features. In particular, we use the embedding $v$ of the step with the highest probability from the previous time instance of the decoder. Following~\cite{sutskever2014sequence}, we further use the attention mechanism over the encoder's states to get another feature ${\bf x}_t^{att}$. In particular:
\vspace{-1mm}
\begin{align}
\alpha_j^t &= \mathrm{softmax}_{j}(v^T \big(W_{att}\,(h^t \| h^j_{enc})\big )\big)\\
{\bf x}_t^{att} &= \sum_j \alpha_j^t h_{enc}^j\\[-8mm]
&\nonumber\
\end{align}
where $W_{att}$, $v$ are learnable parameters. Our full model is visualized in Fig.~\ref{fig:modelVideo}.

%\begin{figure*}[t!]
%\vspace{-2mm}
%\centering
%\begin{tabular}{c}
%\hspace{-6mm}\includegraphics[height=0.215\linewidth,trim=95 5 110 26,clip]{figures/actions_freq.eps}\hspace{0.1mm}
%\includegraphics[height=0.215\linewidth,trim=102 5 108 25,clip]{figures/objects_freq.eps}  \\
%\end{tabular}
%\caption{Frequency of actions ({\bf left}) and objects ({\bf right}) in our \emph{ActivityScripts} dataset.}
%\label{fig:stats}
%\end{figure*}

{\bf Learning and inference}. 
Our goal is to generate programs that are both close to the ground-truth programs in terms of their LCS (longest common subsequence) and are also executable by our renderer. To that end, we train our model in two phases. Firstly, we pre-train the model using cross-entropy loss at each time step of the RNN decoder. We follow the typical training strategy where we make a prediction at each time instance but feed in the ground-truth step to the next time instance. We use the word2vec~\cite{mikolov2013efficient} embeddings for matrices $W_{a}$ and $W_{o}$. 

In the second stage, we  treat program generation as an Reinforcement Learning problem, where the agent is learning a policy that generates steps to compose a program. We follow~\cite{selfcriticalImageCapitoning}, and use policy gradient optimization to train the model, using the greedy policy as the baseline estimator. 
We exploit two different kinds of reward $r(w^s,g)$ for RL training, where $w^s$ denotes the sampled program, and $g$ the ground-truth program. To ensure that the generated program is semantically correct (follows the description/video), we use the normalized LCS  metric (length of the longest common subsequence) between the two programs as our first reward $r_{LCS}(w^s,g)$. The second reward comes from our simulator, and measures whether the generated program is executable or not. This reward, $r_{sim}(w^s)$, is a simple binary value. We carefully balance the total reward as, $r(w^s,g) = r_{LCS}(w^s,g)+0.1\cdot r_{sim}(w^s)$. %In practice, we run a few epochs with the LCS reward alone, then use this model to compute the new baseline, and train a few more epochs with the combined reward. 

So far we did not specify the  input to the RNN encoder. Our model accepts either a language description or a video. %We keep our method general and use the same architecture for both video and text. For text, 

\vspace{-3mm}
\paragraph{Textual Description.} To encode a textual description our RNN encoder gets as input the word2vec~\cite{mikolov2013efficient} embedding of the word in the description at each time instance.

\vspace{-3mm}
\paragraph{Video.} 
To generate programs from videos, we partition each video into 2-second clips and train a model to predict the step at the middle frame. We use DilatedNet to obtain the semantic segmentation of each frame and use the Temporal Relation Network \cite{TRNzhou} with 4-frame relations to predict the embedding of an instruction (action+object+object). We use this embedding to obtain the likelihood of each instruction. The prediction at each clip is used as input to the RNN encoder for program generation.

\begin{table*}[t!]
	\vspace{-1mm}
	\centering
	\addtolength{\tabcolsep}{-0pt}
	\begin{small}
			\scalebox{0.98}{
	\begin{tabular}[t]{|l|c|c|c|c|}
	\hline
	& Action  & Objects        & Steps  & Mean \\
	\hline
	Rand. Retrieval & 8.30\%  & 1.50\%	&   0.51\%  & 3.43\% \\
	Seen homes & 70.32  \%  & 42.14 \% & 23.81 \% & 45.42\%\\
	Unseen homes &31.34\% & 14.55\% & 11.48\% & 19.12\%\\
	All & 46.85\% & 25.76\% & 18.41\% & 30.34\%\\
	
	\hline
	\end{tabular}}
		\hspace{0.3mm}
		\scalebox{0.835}{
		\addtolength{\tabcolsep}{-1pt}
	\begin{tabular}[t]{|l|c|c|c|c||c|}
	\hline
	& Action  & Objects        & Steps  & Mean   & Simulator  \\
	\hline
	Rand. Retrieval & .473 & .079 & .071 & .207 & 100.0\% \\
        MLE 			& .735 & .359 & .341 & .478 & 19.4\% \\
	PG(LCS)  		& .761 & .383 & .364 & .502 & 19.0\% \\
	PG(LCS+Sim)  & .751 & .377 & .358 & .495 & 22.4\% \\
	PG(LCS+Sim) Seen homes &  .851 & .556 & .528 & .645 & 24.6\% \\
	PG(LCS+Sim) Unseen homes & .680 & .250 & .236 & .389 & 20.9\% \\
	\hline
\end{tabular}
}
	\end{small}
	\vspace{-2.5mm}
		\caption{\small {\bf Left:} Accuracy of \emph{video-based action classification} and \emph{action-object-object }(step or instruction in the program) prediction in {\bf 2-sec clips} from our {VirtualHome Activity} dataset. {\bf Right}: \emph{Video-based program generation}. }
\label{table:Triplet classification}
\vspace{-1mm}
\end{table*}

\begin{table*}[t!]
\vspace{-1mm}
	\centering
	\addtolength{\tabcolsep}{0pt}
	\scalebox{0.9}{
	\begin{small}
	\begin{tabular}[t]{|l|c|c|c|c||c|}
		\hline
		Method    & Action  & Objects   & Steps  & Mean  & Simulator (\%) \\
		\hline
		Rand. Sampling&   .226 & .039 & .020 & .095  & 0.6\%\\
		Rand. Retrieval & .473 & .079 & .071 & .207 & 100.0\% \\
		Skipthoughts&   .642 & .272 & .252 & .389 & 100.0\% \\
		% the w2v model with hidden state 100
		% update the w2v attention model with hidden state=100, which has better result
		%W2V Attention&        70.950\%&    59.492\%&       56.155\%&   62.199\%\\
		MLE  &          .777 & .723 & .686 & .729 & 38.6\% \\
		PG(LCS) &    .803 & .766 & .732 & .767 & 35.5\% \\
		PG(LCS+Sim)  &  .806 & .775 & .740 & .774 & 39.8\% \\
		\hline
		\end{tabular}
		\hspace{2mm}
		\scalebox{1.17}{
		\begin{tabular}[t]{|l|c|c|c|c|}
		\hline
		  Method & Action  & Objects        & Steps  & Mean \\
		\hline
		  Rand. Sampling&   .106 & .018 & .004 & .043 \\
		  Rand. Retrieval &	 .320 & .037 & .032 & .130\\
		  Skipthoughts&   .469 & .297 & .266 & .344 \\
		  MLE  &   .497 & .392 & .340 & .410  \\
		 PG(LCS)  &  .522 & .433 & .387 & .447 \\
		\hline
		\end{tabular}
		}
	\end{small}
	}
	\vspace{-2mm}
	\caption{\small{\bf Programs from descr.}: Accuracy on ({\bf left}) \emph{VirtualHome Act.},  and ({\bf right}) \emph{ActivityPrograms}. We compute the length of longest common subsequence between a predicted script and GT and divide by max length of the two programs, mimicking IoU for programs. Since real programs are mainly not executable in our simulator due to the lack of implemented actions, we cannot report the executability metric.}
	\label{table:results}
	\vspace{-2mm}
\end{table*}

\section{Experiments}
\label{sec:results}

%We first analyze our collected \emph{ActivityPrograms} dataset of described real activities and scripts. %, and then evaluate our script generation approach. 
%Since activities in the wild are difficult to execute due to a large diversity of atomic actions they require, we additionally collect a synthetic dataset containing scripts with actions that we can execute in our VirtualHome.

In our experiments we exploit  both of our datasets: \emph{ActivityPrograms} containing descriptions and programs for real activities, and \emph{VirtualHome Activity dataset} that contains synthesized programs, yet natural descriptions to describe them. \emph{VirtualHome Activity dataset} further contains videos animating the programs.

%{\bf ActivityPrograms}. This dataset contains the 1440 programs described in Sec.~\ref{sec:dataset}. Since these programs represent real activities that happen households, they contain significant variability in actions and objects that appear in steps. While our ultimate aim is to be able to animate all these actions in our simulator, our current efforts only support the top 12 actions. We thus create another dataset that contains programs containing only these actions in their steps. 

%{\bf VirtualHome Activity dataset}. We synthesized 5,162  programs using a simple probabilistic grammar, and had each one described in natural language by a human annotator. Although these programs were not given by annotators, they produced reasonable activities, creating a much larger dataset of paired descriptions-programs at a fraction of the cost. We then animated each program in our simulator, and automatically generated ground-truth which allows us to train and evaluate our video models.
%As can be seen from Table~\ref{tab:stats}, descriptions in \emph{VirtualHome Activity} dataset are of comparable length. However, 
%the vocabulary here was biased towards that used in programs. %This also shows later in results. 

%5,162 videos for our \emph{VirtualHome Activity} dataset. Each video has a script and a description, as well as automatically generated ground-truth which allows us to train and evaluate our models. 

\begin{figure*}[t]
\vspace{-2mm}
\begin{center}
\begin{minipage}{0.195\linewidth}
\includegraphics[width=1\linewidth] {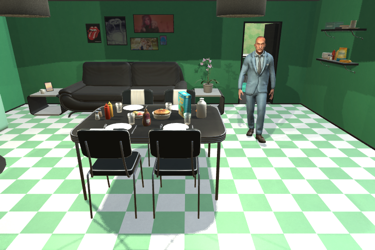}\\[-5.5mm]
{\small\sc\color{magenta}{$\ [${\bf Walk}$]$}}
\end{minipage}
\begin{minipage}{0.195\linewidth}
\includegraphics[width=1\linewidth] {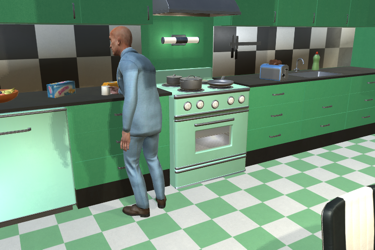} \\[-5.5mm]
{\small\sc\color{magenta}{$\ [${\bf Grab}$]$ $\langle$CUP$\rangle$}}
\end{minipage}
\begin{minipage}{0.195\linewidth}
\vspace{-1mm}
\includegraphics[width=1\linewidth] {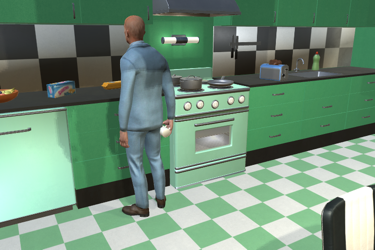}  \\[-5.5mm]
{\small\sc\color{magenta}{$\ \ $}}
\end{minipage}
\begin{minipage}{0.195\linewidth}
\includegraphics[width=1\linewidth] {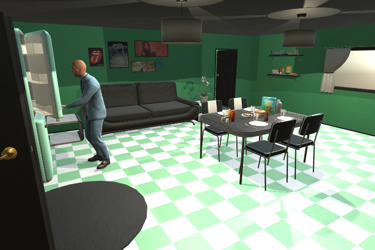} \\[-5.5mm]
{\small\sc\color{magenta}{$\ [${\bf Open}$]$ $\langle$FRIDGE$\rangle$}}
\end{minipage} 
\begin{minipage}{0.195\linewidth}
\includegraphics[width=1\linewidth] {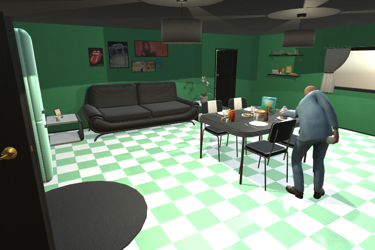} \\[-5.5mm]
{\small\sc\color{magenta}{$\ [${\bf Grab}$]$ $\langle$MILK$\rangle$}}
\end{minipage} 
\\[1.5mm]
{\small {\bf Description:} Get an empty glass. Take milk from refrigerator and open it. Pour milk into glass.}\\[2mm]
\iffalse
\begin{minipage}{0.195\linewidth}
\includegraphics[width=1\linewidth] {figs/results/script2/Action_0120} \\[-5.5mm]
{\small\sc\color{yellow}{$\ [${\bf Walk}$]$}}
\end{minipage} 
\begin{minipage}{0.195\linewidth}
\includegraphics[width=1\linewidth] {figs/results/script2/Action_0380} \\[-5.5mm]
{\small\sc\color{yellow}{$\ [${\bf Sit}$]$ $\langle$CHAIR$\rangle$}}
\end{minipage} 
\begin{minipage}{0.195\linewidth}
\includegraphics[width=1\linewidth] {figs/results/script2/Action_0480} \\[-5.5mm]
{\scriptsize\sc\color{yellow}{$\ [${\bf SwitchOn}$]$ $\langle$COMPUTER$\rangle$}}
\end{minipage} 
\begin{minipage}{0.195\linewidth}
\includegraphics[width=1\linewidth] {figs/results/script2/Action_0515} \\[-5.5mm]
{\scriptsize\sc\color{yellow}{$\ [${\bf SwitchOff}$]$ $\langle$COMPUTER$\rangle$}}
\end{minipage} 
\begin{minipage}{0.195\linewidth}
\vspace{-1mm}
\includegraphics[width=1\linewidth] {figs/results/script2/Action_0544}\\[-5.5mm]
%{\small\sc\color{magenta}{$[${\bf Grab}$]$ $\langle$MILK$\rangle$}}
\end{minipage} 
\\[1.5mm]
{\small {\bf Description:} Turn on computer. Open browser.}\\[2mm]
\fi
\begin{minipage}{0.195\linewidth}
\includegraphics[width=1\linewidth] {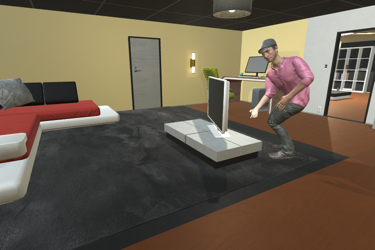}\\[-5.5mm]
{\small\sc\color{cyan}{$\ [${\bf SwitchOn}$]$ $\langle$TV$\rangle$}}
\end{minipage}  
\begin{minipage}{0.195\linewidth}
\vspace{-1mm}
\includegraphics[width=1\linewidth] {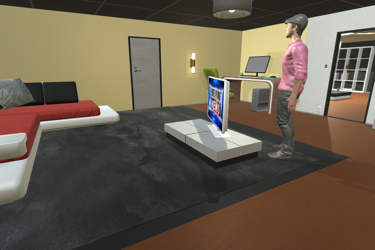} \\[-5.5mm]
%{\small\sc\color{magenta}{$[${\bf Grab}$]$ $\langle$MILK$\rangle$}}
\end{minipage} 
\begin{minipage}{0.195\linewidth}
\includegraphics[width=1\linewidth] {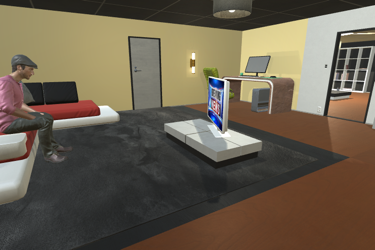} \\[-5.5mm]
{\small\sc\color{cyan}{$\ [${\bf Sit}$]$ $\langle$SOFA$\rangle$}}
\end{minipage} 
\begin{minipage}{0.195\linewidth}
\includegraphics[width=1\linewidth] {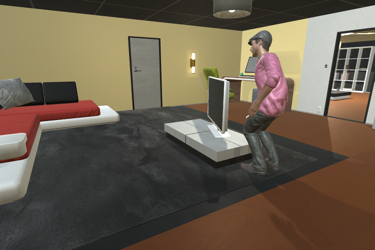} \\[-5.5mm]
{\small\sc\color{cyan}{$\ [${\bf SwitchOff}$]$ $\langle$TV$\rangle$}}
\end{minipage} 
\begin{minipage}{0.195\linewidth}
\includegraphics[width=1\linewidth] {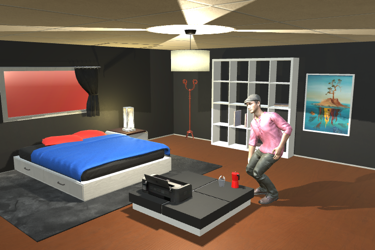} \\[-5.5mm]
{\scriptsize\sc\color{cyan}{$\ [${\bf Put}$]$ $\langle$COFF.-POT$\rangle$ $\langle$TABLE$\rangle$}}
\end{minipage} 
\\[1.5mm]
{\small {\bf Description:} Go watch TV on the couch. Turn the TV off and grab the coffee pot. Put the coffee pot on the table and go turn the light on.}\\[2mm]
\begin{minipage}{0.195\linewidth}
\includegraphics[width=1\linewidth] {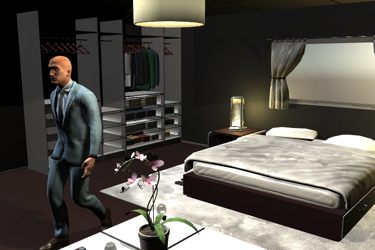}\\[-5.5mm]
{\small\sc\color{green}{$\ [${\bf Walk}$]$}}
\end{minipage}  
\begin{minipage}{0.195\linewidth}
\vspace{-0mm}
\includegraphics[width=1\linewidth] {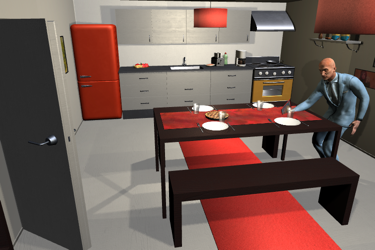} \\[-5.5mm]
{\small\sc\color{green}{$[${\bf Grab}$]$ $\langle$MAGAZINE$\rangle$}}
\end{minipage} 
\begin{minipage}{0.195\linewidth}
\includegraphics[width=1\linewidth] {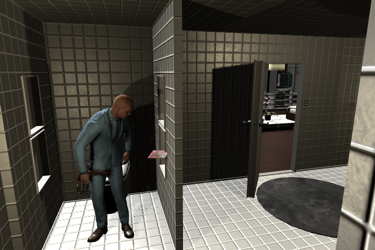} \\[-5.5mm]
{\small\sc\color{green}{$\ [${\bf Sit}$]$ $\langle$TOILET$\rangle$}}
\end{minipage} 
\begin{minipage}{0.195\linewidth}
\includegraphics[width=1\linewidth] {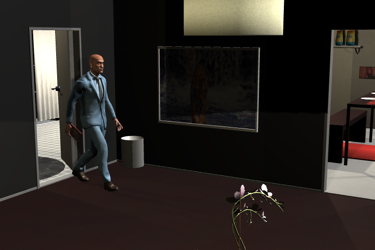} \\[-5.5mm]
{\small\sc\color{green}{$\ [${\bf Walk}$]$}}
\end{minipage} 
\begin{minipage}{0.195\linewidth}
\includegraphics[width=1\linewidth] {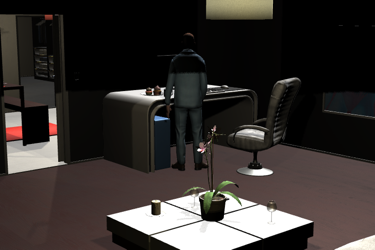} \\[-5.5mm]
{\scriptsize\sc\color{green}{$\ [${\bf Put}$]$ $\langle$MAGAZINE$\rangle$ $\langle$DESK$\rangle$}}
\end{minipage} 
\\[1.5mm]
{\small {\bf Description:} Look at the clock then get the magazine and use the toilet. When done put the magazine on the table.}\\[1.6mm]
\begin{minipage}{0.195\linewidth}
\includegraphics[width=1\linewidth] {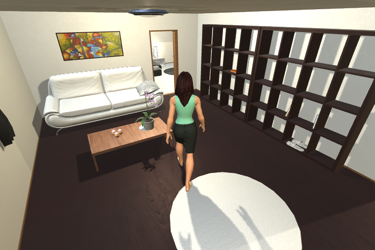}\\[-5.5mm]
{\small\sc\color{red}{$\ [${\bf Walk}$]$}}
\end{minipage}  
\begin{minipage}{0.195\linewidth}
\vspace{-0mm}
\includegraphics[width=1\linewidth] {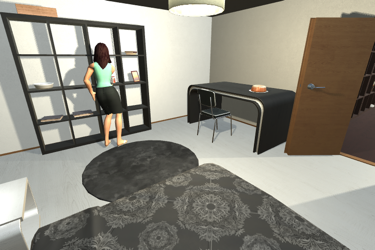} \\[-5.5mm]
{\small\sc\color{red}{$[${\bf Grab}$]$ $\langle$FACE SOAP$\rangle$}}
\end{minipage} 
\begin{minipage}{0.195\linewidth}
\includegraphics[width=1\linewidth] {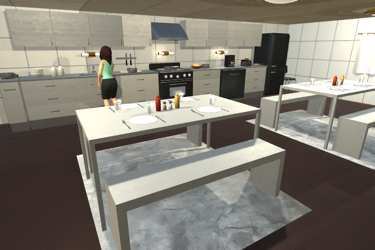} \\[-5.5mm]
{\scriptsize\sc\color{red}{$\ [${\bf Put}$]$ $\langle$F.SOAP$\rangle$ $\langle$COUNTER$\rangle$}}
\end{minipage} 
\begin{minipage}{0.195\linewidth}
\includegraphics[width=1\linewidth] {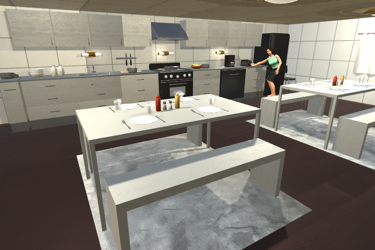} \\[-5.5mm]
{\scriptsize\sc\color{red}{$\ [${\bf SwitchOn}$]$ $\langle$TOASTER$\rangle$}}
\end{minipage} 
\begin{minipage}{0.195\linewidth}
\includegraphics[width=1\linewidth] {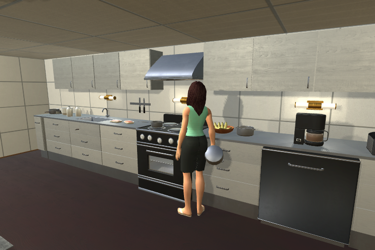} \\[-5.5mm]
{\scriptsize\sc\color{red}{$\ [${\bf Put}$]$ $\langle$POT$\rangle$ $\langle$STOVE$\rangle$}}
\end{minipage} 
\\[1.5mm]
{\small {\bf Description:} Take the face soap to the kitchen counter and place it there. Turn toaster on and then switch it off. Place the pot on the stove.}\\[-1mm]
\caption{\small Our agent executing generated programs from descriptions, in our VirtualHome. Top description is from \emph{ActivityPrograms}, while the rest are from \emph{VirtualHome Activity} dataset. Notice that the top agent uses his left to open the fridge and to grab an object since he already holds an item in his right. There are also some limitations, for example, in row 3 the agent sits on the toilet fully clothed. Furthermore, sometimes the carried item slightly penetrates into the character's body due to imprecisions of the colliders.}
\label{fig:results}
\end{center}
\vspace{-4mm}
\end{figure*}

\vspace{-0.5mm}
\subsection{Step (Instruction) Classification from Video}
\vspace{-0.5mm}
We first evaluate our model for the task of video-based action and action-object-object (step/instruction in the program) classification.
Here, we partition each video in 2-sec clips, and use the clip-based TRN to perform classification. We compute performance as the mean per-class accuracy across all 2-sec clips in \emph{test}. To better understand the generalization properties of the video-based models, we further divide the \emph{test} set into videos recorded in \emph{homes seen} at train time, and videos in \emph{homes not seen} at train time.  We report the results in Table~\ref{table:Triplet classification} (left). To set the lower bound, we also report a simple \emph{random retrieval} baseline, in which a step is randomly retrieved from the training set. We can see that our model performs significantly better. However, a large number of actions and objects of interest, makes the prediction task challenging for the model. 

\vspace{-0.5mm}
\subsection{Program Generation}
\vspace{-0.5mm}

\begin{figure}[htb]
\vspace{-4mm}
\centering
\begin{tabular}{c}
\includegraphics[width=0.75\linewidth,trim=0 200 0 210,clip]{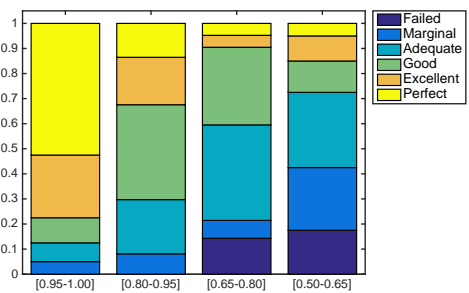}
\end{tabular}
\vspace{-4mm}
\caption{\small Human evaluation of our agent executing described activities via program generation from text: x axis shows the program prediction accuracy, y axis is the human score.}
\label{fig:human}
\vspace{-3mm}
\end{figure}

We now evaluate the task of program generation.  

{\bf Metrics}. We evaluate program induction using a measure similar to IOU. We compute the longest common subsequence between a GT and a predicted program, where we allow gaps between the matching steps, but require their order to be correct. We obtain accuracy as the length of the subseq. divided by the max of the two programs' lengths.  We also compute accuracies for \emph{actions} and \emph{objects} alone. Since LCS does not measure whether the program is valid, we report another metrics that computes the percentage of generated programs that are executable in our simulator. 
%\textcolor{red}{We also report, for the VirtualHome scripts, the percentage of the generated scripts that are executable according to our parser (PythonSim) and the ones executable in the VirtualHome simulator. \textcolor{red}{should specify the difference between the 2}} %, to analyze the model in more detail.

{\bf Language-based prediction}. Since we have descriptions for all activities, we first evaluate how well our model translates natural language descriptions into programs.
We report results on \emph{ActivityPrograms} (real activities), as well as on \emph{VirtualHome Activity} datasets (where we first only consider descriptions, not videos). %For the former, we automatically generate scripts restricted to our implemented atomic actions, using a simple probabilistic grammar. We then show the scripts to annotators and ask them to describe it in natural language, as if they were to describe instructions to a friend. 
%As can be seen from Table~\ref{tab:stats}, the descriptions for the \emph{VirtualHome Activity} dataset are of comparable length. However, the 
%since these descriptions were obtained by asking annotators to describe scripts in natural language, the vocabulary was biased towards that used in scripts. This also shows in results. 
We compare our models to four baselines: 1) \emph{random sampling}, where we randomly pick both an action for each step and its arguments, 2) \emph{random retrieval}, where we randomly pick a program from the training set, 3) \emph{skipthoughts}, where we embed the description using~\cite{skipthoughts,ZhuICCV15}, retrieve the closest description from training set and take its program, 4) our model trained with MLE (no RL). 
%\textcolor{red}{Seq2seq with cross-entropy loss, where the seq2seq model with attention described in Figure \ref{fig:modelVideo}}.
Table~\ref{table:results} provides the results. We can see that our model outperforms all baselines on both datasets. Our RL model that exploits LCS reward outperforms the MLE model on both metrics (LCS and executability). Our model that uses both rewards slightly decreases the LCS score, but significantly improves the executability metrics. %Notice that scripts from \emph{VirtualHome Activity} dataset are significantly easier than the ``real'' ones. %This indicates that if an agent is given a more detailed description, it is likely to execute the action correctly. 
%This means that if the activities were to be described to an artificial agent in more plain language, 

{\bf Video-based prediction.} We also report results on the most challenging task of video-based program generation. %Here, we train our base model using the ground-truth steps of the 2 second clips and fine-tune it using the predicitons of our step classification model. We further use the approach described in \ref{} and train 2 models with LCS and LCS + Sim rewards.} .
The results are shown in Table~\ref{table:Triplet classification} (right). One can observe that RL training with LCS reward improves the overall accuracy over the MLE model (the generated programs are more meaningful given the description/video), however its executability score decreases. This is expected: MLE model typically generates shorter programs, which are thus more likely to be executable (an empty program is always executable). A careful balance of both metrics is necessary.  RL with both the LCS and the simulator reward improves both LCS and the executability metrics over the LCS-only model.

{\bf Executing programs in VirtualHome.} In Fig.~\ref{fig:results} we show a few examples of our agent executing programs generated from natural descriptions. To understand the quality of our simulator as well as the plausibility of our program evaluation metrics, we perform a human study. We randomly selected 10 examples per level of performance: (a) $[0.95-1]$, (b) $[0.8-0.95]$, (c) $[0.65-0.8]$, and (d)  $[0.5-0.65]$.  For each example we had 5 AMT workers judge the quality of the performed activity in our simulator, given its language description. %We showed the description of the action, and asked the annotator to judge how well the agent executed the described action. 
Results are shown in Fig.~\ref{fig:human}. One can notice agreement between our metrics and human scores. Generally, at perfect performance the simulations got high human scores, however, there are examples where this was not the case. This may be due to imperfect animation, an indication that further improvements to our simulator are possible.
%Additional results are in the supplementary material.

{\bf Implications.} The high performance of text-based activity animation opens exciting possibilities for the future. %It indicates that novel activity videos could be reliably generated by human annotators, describing them in natural language. 
It would allow us to replace the more rigid program synthesis that we used to create our dataset, by having annotators create these animations directly via natural language or crowd-sourcing scripts from existing text corpora.

\vspace{-1mm}
\section{Conclusion}
\label{sec:conc}
\vspace{-1mm}

We collected a large knowledge base of how-to for household activities specifically aimed for robots. Our dataset contains natural language descriptions of activities as well as \emph{programs}, a formal symbolic representation of activities in the form of a sequence of steps. What makes these programs unique is that they contain \emph{all} the steps necessary to perform an activity. We further introduced \emph{VirtualHome}, a 3D simulator of household activities, which we  %We first collected a dataset of descriptions and scripts for activities. Top most frequent actions and interactions were then implemented in our simulator. 
used to create a large video activity dataset  with rich ground-truth. We proposed a simple model that infers a program from either a video or a textual description,
allowing robots to be ``driven'' by naive users via natural language or video demonstration. 
We  showed examples of agents performing these programs in our simulator. There are many exciting avenues going forward, for example, training agents to perform tasks from visual observation alone using RL techniques.
%Our agent is given a description of a composite action in natural language is tasked to execute the action in the enviroment. We proposed a neural encoder-decoder model that translates a description of the action into a symbolic script which can be executed in our VirtualHome, potentially showcasing how a robotic platform could follow instructions from humans via natural language. Going forward, there are many exciting avenues for our VirtualHome, for example, such as serving as a testbed for  reinforcement learning techniques.

\fontsize{7,5pt}{9.2pt}\selectfont \textbf{Acknowledgements:} We acknowledge partial support from "La Caixa" Fellowship, NSERC COHESA NETGP485577-15, Samsung, DARPA Explainable AI (XAI) program and IARPA D17PC00341. The U.S. Government is authorized to reproduce and distribute reprints for Governmental purposes notwithstanding any copyright annotation thereon. Disclaimer: The views and conclusions contained herein are those of the authors and should not be interpreted as necessarily representing the official policies or endorsements, either expressed or implied, of IARPA, DOI/IBC, or the U.S. Government. We also gratefully acknowledge NVIDIA for donating several GPUs used in this research.

{\small
\bibliographystyle{ieee}
\bibliography{egbib}
}

\end{document}